\documentclass[11pt,a4paper]{article}
\usepackage[utf8]{inputenc}
\usepackage[hyperref]{acl2020}
\usepackage{times}
\usepackage{latexsym}
\usepackage{amsmath,amssymb,amsthm,amsfonts}
\usepackage{graphicx}
\usepackage{blkarray}
\usepackage{booktabs}
\usepackage{subcaption}
\usepackage{url}
\usepackage{wrapfig}
\usepackage{microtype}
\usepackage{color, colortbl}
\usepackage{tikz}
\usepackage{pgfplots}
\usepackage[arabic,farsi,vietnamese,english]{babel}
\usepackage[T5,T1]{fontenc}
\usepackage[colorinlistoftodos,textsize=footnotesize,textwidth=2cm,color=yellow!15, linecolor=yellow]{todonotes}

\usetikzlibrary{arrows,arrows.meta,matrix,calc,positioning,decorations.pathreplacing}

\newcommand{\key}{\textbf}

\makeatletter
\renewcommand*{\p@section}{\S}
\renewcommand*{\p@subsection}{\S}

\aclfinalcopy 

\title{Predicting cross-linguistic adjective order with information gain}

\author{
  William Dyer\\
  Oracle Corporation\\
  \texttt{william.dyer@oracle.com}
  \\\And
  Richard Futrell\\
  University of California, Irvine\\
  \texttt{rfutrell@uci.edu}
  \\\AND
  Zoey Liu\\
  Boston College\\
  \texttt{ying.liu.5@bc.edu}
  \\\And
  Gregory Scontras\\
  University of California, Irvine\\
  \texttt{g.scontras@uci.edu}
}

\date{}

\begin{document}
\maketitle

\begin{abstract}
Languages vary in their placement of multiple adjectives before, after, or surrounding the noun, but they typically exhibit strong intra-language tendencies on the relative order of those adjectives (e.g., the preference for `big blue box' in English, `grande boîte bleue' in French, and `al\d{s}und\={u}q al’azraq alkab\={\i}r' in Arabic). We advance a new quantitative account of adjective order across typologically-distinct languages based on maximizing information gain. Our model addresses the left-right asymmetry of French-type ANA sequences with the same approach as AAN and NAA orderings, without appeal to other mechanisms. We find that, across 32 languages, the preferred order of adjectives largely mirrors an efficient algorithm of maximizing information gain.
\end{abstract}

\section{Introduction}
Languages that allow multiple sequential adjective modifiers tend to exhibit strong tendencies on the relative order of adjectives, as in  `big blue box' vs.~`blue big box' in English \citep{dixon1982where}. 
To date, most of the research on adjective ordering has focused on preferences in pre-nominal languages like English where adjectives precede the modified noun \citep{futrell2020what}, or in post-nominal languages like Arabic where adjectives follow the noun \citep{kachakechescontras2020}. 
This research usually posits a metric, such as information locality \citep{futrell2020lossy} 
or subjectivity \citep{scontras2017subjectivity}, which governs the preferred distance between a noun and its adjectives. Because these theories predict only the relative linear \emph{distance} between noun and adjective, they cannot be straightforwardly applied to mixed languages like French, where adjectives regularly appear both before and after the modified noun, at least not without added assumptions about hierarchical distance \citep{cinque_evidence_1994}. Instead, these mixed languages are often modeled with constraints on which adjective classes or functions can appear before or after a noun \citep{cinque_syntax_2010,fox2012predicting}. 

 
Traditional accounts of adjective ordering in the linguistics literature often assume a \key{tree structure} 
in which the target measure is the hierarchical distance from noun (N) to adjective (A). 
According to syntactic accounts, ordering regularities are predicted by a universal hierarchy of lexical semantic classes (e.g., color adjectives are hierarchically closer to the modified noun than size adjectives; \citealp{cinque_evidence_1994,scott_stacked_2002}). Alternative accounts use aspects of adjective meaning to predict adjective order, making appeal to notions like `inherentness' \citep{whorf_grammatical_1945} or `definiteness of denotation' \citep{martin1969}. Recently, \citet{scontras2017subjectivity} provide experimental evidence that their synthesis of semantic predictors into a continuum based on subjectivity reliably predicts ordering preference in English; followup studies have found subjectivity to be a reliable predictor in other languages as well (Tagalog: \citealp{samontescontras2019}; Mandarin: \citealp{shiscontras2020}; Arabic: \citealp{kachakechescontras2020}; Spanish: \citealp{rosalesscontras2019,scontrasetal2020}). Explanations for the role of subjectivity in adjective ordering show how subjectivity-based orderings are more \key{efficient} than alternative orderings, thereby maximizing communicative success \citep{simonic2018,hahn2018information,frankeetal2019,scontrasetal2019}.

Other efficiency-based approaches to adjective order quantify efficiency with information-theoretic measures of word distributions such as surprisal or entropy \citep{cover2006elements,levy2008expectation}. Models in this vein have a long conceptual history in the field, originating with the idea that semantic closeness between words is reflected in syntactic closeness in a surface realization \citep{sweet_new_1900,jespersen_language_1922,behaghel_deutsche_1932}. Modern quantitative incarnations include integration cost \citep{dyer2017minimizing} and information locality \citep{futrell2020lossy}, both generalizations of the widely-accepted principle of dependency distance minimization \citep{liu_dependency_2017,temperley2018minimizing}.

Crucially, while previous approaches are able to model symmetrical structures within the noun phrase, as in the mirror-image $A_1A_2N$ orders of English and the $NA_2A_1$ orders of Arabic, a hierarchical approach cannot model the \key{left--right asymmetry} of Romance $A_1NA_2$ without an appeal to other mechanisms. 

We advance an information-theoretic factor that predicts adjective ordering across the three typological `templates' of adjective order\textemdash pre (AAN), mixed (ANA), and post (NAA)\textemdash based on \key{information gain} (IG), a measure of the reduction in uncertainty attained by transforming a dataset. IG is used in machine learning for ordering the nodes of a decision tree \cite{quinlan_induction_1986,norouzi_efficient_2015}, where nodes are most often ordered in a greedy fashion such that the information gain of each node is maximized. By analogy, we view the noun phrase as a decision tree for reducing a listener's uncertainty about a speaker's intended meaning. Each word acts as a node in the decision tree; preferred adjective orders thus reflect an efficient ordering of nodes.

\section{Empirical background}


Empirical investigations of adjective ordering have focused on the cross-linguistic stability of these preferences across a host of unrelated languages (e.g., \citealp{dixon1982where,hetzron1978on,sproat_cross-linguistic_1991}).
For example, where English speakers prefer `big blue box' to `blue big box', Mandarin speakers similarly prefer \emph{d\`{a}-de l\'{a}n-de xi\={a}ng-zi} `big blue box' to \emph{l\'{a}n-de d\`{a}-de xi\={a}ng-zi} `blue big box' \citep{shiscontras2020}. In post-nominal languages, we find the mirror-image of the English pattern, such that adjectives that are preferred closer to the noun in pre-nominal languages are also preferred closer to the noun in post-nominal languages.\footnote{Celtic languages have been claimed to be an exception to this trend \citep{sproat_cross-linguistic_1991}, though our own investigations into Irish suggest that it behaves like other post-nominal languages, at least with respect to information gain.} For example, speakers of Arabic prefer \emph{al\d{s}und\={u}q al’azraq alkab\={\i}r} `the box blue big' to \emph{al\d{s}und\={u}q alkab\={\i}r al’azraq} `the box big blue'.

In support of the cross-linguistic stability of adjective ordering preferences, \citet{leungetal2020} present a latent-variable model capable of accurately predicting adjective order in 24 languages from seven different language families, achieving a mean accuracy of 78.9\% on an average of 1335 sequences per language. Importantly, the model succeeds even when the training and testing languages are different, thus demonstrating that different languages rely on similar preferences. However, \citeauthor{leungetal2020}'s study was limited to AAN and NAA templates. There has been very little corpus-based empirical work on ordering preferences in the mixed ANA template, where adjectives both precede and follow the modified noun.\footnote{We note two studies that have examined adjective orders in five Romance languages from a quantitative perspective:~\citet{gulordava-etal-2015-dependency} and~\citet{gulordava-merlo-2015-structural}. Different from our work, their studies looked at noun phrases with only one adjective phrase dependent.}

While \citet{leungetal2020} learn adjective order by training on observed adjective pairs, an alternate strategy is to posit one or more a priori metrics as an underlying motivation for adjective order (e.g., \citealp{malouf2000order}, in part). This approach allows for the study of why adjective orders might have come about. To that end, \citet{futrell2020what} report an accuracy of 72.3\% for English triples based on a combination of subjectivity and information-theoretic measures derived from the distribution of adjectives and nouns.


To our knowledge, the current study is the first attempt at predicting adjective order across all three templates, with an eye not only to raw accuracy, but in hopes of illuminating the functional pressures which might contribute to word ordering preferences in general.

\section{Information gain}

\subsection{Picture of communication}
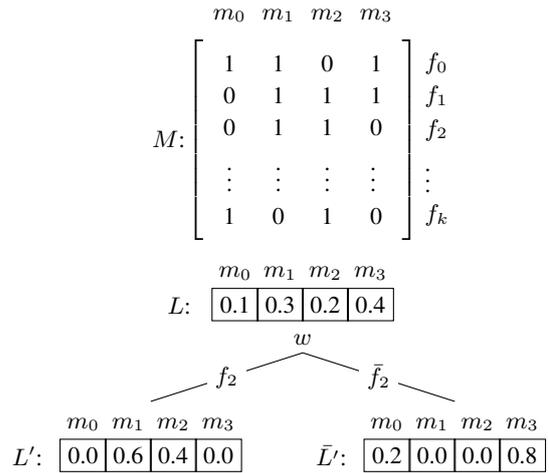
\begin{figure}
\centering \small
\begin{tikzpicture}[baseline]

\begin{scope}[yshift=0cm]
\matrix(M)[matrix of nodes, column 4/.style={anchor=base west}, left delimiter={[},right delimiter={]}, column sep=0.275cm]
{1 &1 &0 &1\\
0 &1 &1 &1\\
0 &1 &1 &0\\
\vdots &\vdots &\vdots &\vdots\\
1 &0 &1 &0\\
};
\end{scope}
\matrix (M2) [matrix of nodes, right= 0.4em of M, column 1/.style={anchor=base west}, row sep=-0.05cm]
{$f_0$ \\$f_1$ \\$f_2$ \\$\vdots$ \\$f_k$ \\};
\matrix (M3) [matrix of nodes, above= 0.0em of M]
{\node(){$m_0$}; &\node(){$m_1$}; &\node(){$m_2$}; &\node(){$m_3$};\\};
\node(Mlabel) at ([xshift=-0.5cm]M.west) {$M$:};

\begin{scope}[yshift=-2cm]
      \matrix(L)[matrix of nodes,
          row 1/.style={nodes={draw=none, minimum width=0.25cm}},
          nodes={draw}, column sep=-0.05cm]
      {
          $m_0$ & $m_1$ & $m_2$ & $m_3$\\
          0.1 &0.3 &0.2 &0.4\\
      };
\end{scope}
\node(Llabel) at ([xshift=-0.3cm, yshift=-0.2cm]L.west) {$L$:};

\begin{scope}[xshift=-2cm, yshift=-4cm] \small
      \matrix(Lp)[matrix of nodes,
          row 1/.style={nodes={draw=none, minimum width=0.3cm}},
          nodes={draw}, column sep=-0.05cm]
      {
          $m_0$ & $m_1$ & $m_2$ & $m_3$\\
          0.0 &0.6 &0.4 &0.0\\
      };
\end{scope}
\node(Lplabel) at ([xshift=-0.3cm, yshift=-0.2cm]Lp.west) {$L^\prime$:};

\begin{scope}[xshift=2cm, yshift=-4cm] \small
      \matrix(bLp)[matrix of nodes,
          row 1/.style={nodes={draw=none, minimum width=0.3cm}},
          nodes={draw}, column sep=-0.05cm]
      {
          $m_0$ & $m_1$ & $m_2$ & $m_3$\\
          0.2 &0.0 &0.0 &0.8\\
      };
\end{scope}
\node(Lplabel) at ([xshift=-0.3cm,yshift=-0.2cm]bLp.west) {$\bar{L^\prime}$:};

\node(w) at (0, -2.65) {$w$};

\draw (w.south) --node[fill=white] {$f_2$} (Lp.north);
\draw (w.south) --node[fill=white] {$\bar{f_2}$} (bLp.north);

\end{tikzpicture}
\caption{A toy universe composed of four feature vectors $m$ defined by $k$ binary features $f$ and an associated probability distribution $L$. Partitioning $L$ on $f_2$ yields $L^\prime$, the probability distribution of the feature vectors containing a 1 for $f_2$, viz. $m_1$ and $m_2$, as well as $\bar{L^\prime}$, the distribution of feature vectors containing a 0 for $f_2$, or $\bar{f_2}$.}
\label{fig:toy}
\end{figure}

We assume that a speaker is trying to communicate a \key{meaning} to a listener, with a meaning represented as a binary vector, where each dimension of the vector corresponds to a feature. Multiple features can be true simultaneously. For example, a speaker might have in mind a vector like $m_1 = [ 1 1 1 \ldots 0]$ in Figure \ref{fig:toy},
where the vector has value 1 in the dimensions for `is-big' ($f_0$), `is-grey' ($f_1$), and `is-elephant' ($f_2$), and 0 for all other features. A meaning of this sort would be conveyed by the noun phrase `big grey elephant'. 
We call $m$ a \key{feature vector} and the set of feature vectors $M$.

The listener does not know which meaning $m$ the speaker has in mind; the listener's state of uncertainty can be represented as a probability distribution over all possible feature vectors, $P(m)$, corresponding to the prior probability of encountering a given feature vector. We call this distribution the \key{listener distribution} $L$.

By conveying information, each word in a sequence causes a change in the listener's prior distribution. 
Suppose as in Figure \ref{fig:toy} that a listener starts with probability distribution $L$, then hears a word $w$ conveying a feature ($f_2$), resulting in the new distribution $L^\prime$. The amount of change from $L$ to $L^\prime$ is properly measured using the Kullback–Leibler (KL) divergence $D_{\text{KL}}[L^\prime||L]$ \citep{cover2006elements}. Therefore, the divergence $D_{\text{KL}}[L^\prime||L]$ measures the amount of information about meaning conveyed by the word.

Another measure of the change induced by a word is the information gain, an extension of KL divergence to include the notion of negative evidence. Let $\bar{L^\prime}$ represent the listener's probability distribution over feature vectors 
conditional on the negation of $w$. By taking a weighted sum of the positive and negative KL divergence, we recover \key{information gain} \citep{quinlan_induction_1986}:
\begin{equation}
 \label{eq:ig-kl}
    \text{IG} = \frac{|L^\prime|}{|L|} D_{\text{KL}}[L^\prime || L] + \frac{|\bar{L}^\prime|}{|L|}D_{\text{KL}}[\bar{L^\prime}||L],
\end{equation}
where $|L|$ indicates the number of elements in the support of $L$ with non-zero probability.
Information gain represents the information conveyed by a word and also the information conveyed by its negation. 

Below, we discuss how information gain relates to other information-theoretic quantities, and why it is useful for us for predicting adjective order across typological templates.

\subsection{Relationship to other quantities}\label{ssec:relationship}

Our IG quantity in Eq.~\ref{eq:ig-kl} is drawn from the ID3 algorithm for generating decision trees \citep{quinlan_induction_1986}. The goal of ID3 is to produce a classifier for some random variable (call it $L$) which works by successively evaluating some set of binary features in some order. The optimal order of these features is given by greedily maximizing information gain, where information gain for a feature $f$ is a measure of how much the entropy of $L$ is decreased by partitioning the dataset into positive and negative subsets based on whether $f$ is present or absent. Our application of information gain to word order comes from treating each word as a binary indicator for the presence or absence of the associated feature, and then applying the ID3 algorithm to determine the optimal order of these features.





The first term of Eq.~\ref{eq:ig-kl}, the divergence $D_{\text{KL}}[L^\prime||L]$, measures the amount of information about $L$ conveyed by the word $w$ and has been the subject of a great deal of study in psycholinguistics. In particular, \citet{levy2008expectation} shows that if the word $w$ and the context $c$ can be reconstructed perfectly from the updated belief state $L^\prime$, then the amount of information conveyed by $w$ reduces to nothing other than the \key{surprisal} of word $w$ in context $c$:
\begin{equation}
    D_{\text{KL}}[L^\prime || L] = -\log p(w|c).
\end{equation}
Importantly for our purposes, the positive evidence term $D_{\text{KL}}[L^\prime||L]$ alone is unlikely to make useful predictions about cross-linguistic word-order preferences, because surprisal is invariant to reversal of word order across a language as a whole \citep{levy2005probabilistic,futrell2019information}: the same surprisal values would be measured for any given language and a language with all the same sentences in reverse order. Therefore, these metrics are unable to predict any a priori asymmetries in word-order preferences between pre- and post-nominal positions.

\subsection{Negative evidence}

The new feature of information gain, which has not been presented in previous information-theoretic models of language, is the negative evidence term in $D_{\text{KL}}[\bar{L}^\prime||L]$, indicating the change in the listener's belief about $L$ given the negation of the features indicated by word $w$, a quantity related to extropy \cite{lad2015extropy}. For example, consider \emph{académie militaire} `military academy' in French. Let $L$ represent a listener's belief state after having heard the noun \textit{académie} `academy'. Upon hearing the adjective \textit{militaire} `military', $L$ is partitioned into $L^\prime$\textemdash the portion of $L$ in which \textit{militaire} is a feature\textemdash and $\bar{L^\prime}$, the portion of $L$ in which \textit{militaire} is not a feature. Put another way, $\bar{L^\prime}$ is the probability distribution over non-military academies.

The negative evidence portion of information gain is of primary interest to us because it breaks the symmetry to word-order reversal that we would have if we used the positive evidence term alone. Therefore, information gain can predict left--right asymmetrical word-order preferences such as the order of adjectives in ANA templates; it also maps onto a well-known decision rule for the ordering of trees.

\section{Methodology}

\subsection{Data}
\label{ssec:data}
Our study relies on two types of source data, both extracted from the CoNLL 2017 Shared Task: Multilingual Parsing from Raw Text to Universal Dependencies \citep{ginter2017, zeman_conll_2017} a set of Common Crawl and Wikipedia text data across a variety of languages, automatically parsed according to the Universal Dependencies scheme with UDPipe \citep{straka_tokenizing_2017}. First, we extract \key{noun phrases} (NPs) containing at least one adjective as the source of feature vectors (\ref{ssec:feature-vecs}). Second, we extract \key{triples}, instances of a noun and two dependent adjectives in any order, where the three words are sequential in the surface order and neither the noun nor the adjectives have any other dependents. 

We restrict triples in this way to minimize the effect that other dependents might have on order preferences. For example, while single-word adjectives tend to precede the noun in English, as in `the \textit{nice} people', adjectives in larger right-branching phrases often follow: `the people \textit{nice to us}' \citep{matthews_positions_2014}, a trend also seen in Romance \citep{gulordava-etal-2015-dependency, gulordava-merlo-2015-structural}. Similarly, conjunctions have been shown to weaken or neutralize preferences \citep{fox2012predicting,
rosalesscontras2019, scontrasetal2020}.

NPs and triples extracted from the Wikipedia dumps are used to generate feature vectors and to train our regression (\ref{ssec:evaluation}). We use triples from the Common Crawl dumps to perform hold-out accuracy testing.

\subsection{Normalization}
Because our source data are extracted from dumps of automatically-parsed text, they contain a large amount of noise, such as incorrectly assigned syntactic categories, HTML, nonstandard orthography, and so on. To combat this noise, we extract all lemmas marked as ADJ and NOUN in all Universal Dependencies (UD) v2.7 corpora \citep{ud2.7} for a given language\textemdash the idea being that the UD corpora are of higher quality\textemdash and include only NPs and triples in which the adjectives and nouns are in the UD lists. All characters are case-normalized, where applicable.

\subsection{Feature vectors}\label{ssec:feature-vecs}
Each NP attested in the Wikipedia corpus for a given language corresponds to a feature vector with value 1 in the dimension associated with each adjective or noun lemma. For example, an NP such as ``the best room available'' generates a vector containing 1 for `is-available', `is-best', and `is-room'.

The relative count of each NP in the Wikipedia corpus yields a probability distribution on feature vectors. It is this distribution which is transformed by partitioning on each lemma in a triple. 


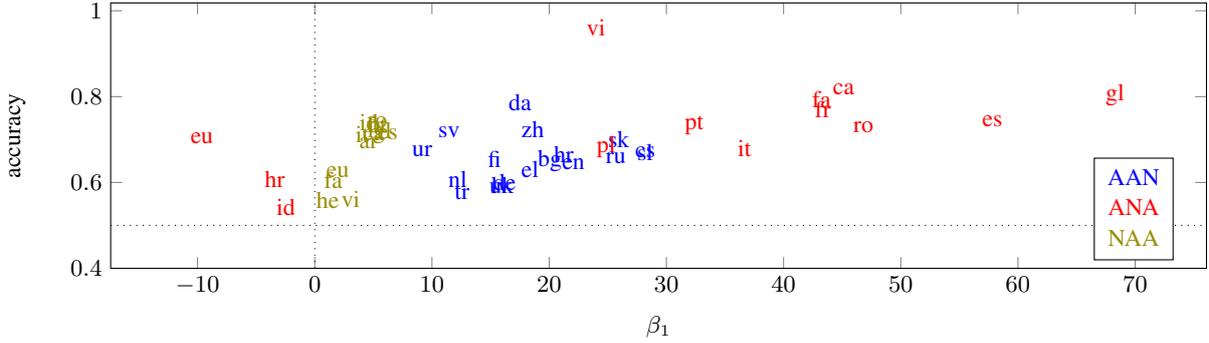
\begin{figure*}
\centering
\begin{tikzpicture}\footnotesize
\begin{axis}[
    ylabel=accuracy,
    xlabel=$\beta_1$,
    legend pos=south east,
    width=\textwidth,
    height=5.1cm,
    ymin=0.4,
    legend style={empty legend}
    ]
    \addplot[
            visualization depends on={value \thisrow{label} \as \Label},
    scatter/classes={AAN={mark=text, text mark=\Label, blue}, ANA={mark=text, text mark=\Label, red}, NAA={mark=text, text mark=\Label, olive}},
        scatter, draw=none, scatter src=explicit symbolic
    ] table [meta=class] {
x y class label
20.058 0.650 AAN bg
28.207 0.671 AAN cs
17.506 0.786 AAN da
16.210 0.601 AAN de
18.383 0.631 AAN el
22.076 0.643 AAN en
15.342 0.655 AAN fi
21.246 0.666 AAN hr
15.826 0.594 AAN lv
12.201 0.609 AAN nl
25.697 0.658 AAN ru
25.935 0.700 AAN sk
28.192 0.670 AAN sl
11.462 0.717 AAN sv
12.579 0.576 AAN tr
15.949 0.593 AAN uk
9.170 0.673 AAN ur
18.604 0.724 AAN zh
45.135 0.818 ANA ca
57.813 0.744 ANA es
-9.623 0.703 ANA eu
43.242 0.794 ANA fa
43.349 0.771 ANA fr
68.290 0.805 ANA gl
-3.411 0.608 ANA hr
-2.462 0.543 ANA id
36.658 0.681 ANA it
24.873 0.684 ANA pl
32.374 0.734 ANA pt
46.823 0.730 ANA ro
24.013 0.962 ANA vi
4.595 0.693 NAA ar
5.024 0.710 NAA ca
6.214 0.713 NAA es
1.957 0.626 NAA eu
1.583 0.605 NAA fa
5.143 0.737 NAA fr
5.776 0.716 NAA gl
1.115 0.558 NAA he
4.631 0.740 NAA id
4.057 0.713 NAA it
5.329 0.726 NAA pt
5.333 0.742 NAA ro
3.068 0.561 NAA vi
    };

\draw[dotted] (axis cs:0,\pgfkeysvalueof{/pgfplots/ymin}) -- (axis cs:0,\pgfkeysvalueof{/pgfplots/ymax});
\draw[dotted] (axis cs:\pgfkeysvalueof{/pgfplots/xmin},0.5) -- (axis cs:\pgfkeysvalueof{/pgfplots/xmax},0.5);
    \addlegendentry[blue]{AAN}
    \addlegendentry[red]{ANA}
    \addlegendentry[olive]{NAA}
\end{axis}
\end{tikzpicture}
\caption{Plot of accuracy and $\beta_1$ coefficient, categorized by template type.}
\label{fig:plot}

\end{figure*}

\subsection{Evaluation}\label{ssec:evaluation}
For a given typological template (AAN, ANA, or NAA) there are two competing variants; our tasks are to (i) predict which of the variants will be attested in a corpus and (ii) show a cross-linguistic consistency in how that prediction comes about.

Because we are limiting our study to the two competing variants within each template, the position of the noun is invariant, leaving only the relative order of the two adjectives to determine the order of a triple. Our problem thus reduces to whether the information gain of the first linear adjective is greater than that of the second. 

In the case of AAN and ANA triples, the IG of each adjective is calculated by partitioning the entire set of feature vectors $L$ on each of the two adjectives. In the case of NAA triples, however, IG is calculated by partitioning only those feature vectors which `survive' the initial partition by the noun, and are therefore part of $L^\prime$. Thus we calculate $\text{IG}(L, a)$ before the noun and $\text{IG}(L^\prime, a)$ after.

Rather than simply implement the ID3 algorithm and choose adjectives based on their raw information gain, we train a logistic regression to predict surface orders based on the \emph{difference} of IG between the attested first and second adjective, a method previously used by \citet{morgan2016abstract} and \citet{futrell2020what}. The benefits of this approach are two-fold: we are able to account for bias in the distribution of adjectival IGs, and we can more easily deconstruct how strong information gain is as a predictor of adjective order.

Within each template, for each attested triple $\tau$, let $\pi_1$ be the lexicographically-sorted first permutation of $\tau$ and $\pi_2$ be the second, with $\alpha_1$ being the first linear adjective in $\pi_1$ and $\alpha_2$ being the first linear adjective in $\pi_2$. Our independent variable $p$ is whether $\pi_1$ is attested in the corpus, and our dependent variable is the difference between the information gain of $\alpha_1$ and $\alpha_2$. We train the coefficients $\beta_0$ and $\beta_1$ in a logistic regression of the form
\begin{equation}
\label{eq:regression1}
\begin{split}
p &= 
\begin{cases}
     1, & \text{if}\ \pi_1 \text{ is attested} \\
     0, & \text{if}\ \pi_2 \text{ is attested} \\
\end{cases}\\
\log\frac{p}{1 - p} &\sim \beta_0 + \beta_1 [\text{IG}(\alpha_1) - \text{IG}(\alpha_2)].
\end{split}
\end{equation}
A positive value for $\beta_1$ tells us that permutations in which the larger-IG adjective is placed first tend to be attested. The value of $\beta_0$ tells us whether there is a generalized bias towards a positive or negative $\text{IG}(\pi_1) - \text{IG}(\pi_2)$. The accuracy we achieve by running the logistic regression on held-out testing data tells us the effectiveness of an IG-based algorithm at predicting adjective order.

\subsection{Reporting results}
We report results for languages from which at least 5k triples could be analyzed, and for templates representing at least 10\% of a language's triples in UD corpora. The count of analyzable triples for each language is a product of those available in the 2017 CoNLL Shared Task, those with sufficiently large UD v2.7 corpora, and those that meet our extraction requirements (\ref{ssec:data}).

Because we are interested in exploring a cross-linguistic predictor of adjective order, we report macro-average accuracies and $\beta_1$ coefficients. That is, each language's accuracy and coefficient are calculated independently and are then averaged. We report both type- and token-accuracy, using the latter in our analysis based on the intuition that the preference for the order of a commonly-occurring triple is stronger than a more rare one.

\section{Results}
We extracted and analyzed at least 5k triples from 32 languages across a variety of families. Because some languages contain triples in two typological templates, we report results for 44 sets of triples. Table \ref{tab:results} reports language-specific results and means for each template, including $n$ triples analyzed, regression coefficient $\beta_1$ and $P$-value, token and type accuracy, and 95\% confidence intervals. Figure~\ref{fig:plot} shows a plot of accuracy and $\beta_1$ coefficient for each language, categorized by template.

\begin{table*}
\centering \def\arraystretch{0.935} \setlength{\tabcolsep}{10pt}
\begin{tabular}[t]{@{} p{4cm} l r r r r r @{}}
\toprule
\textbf{AAN} &language &$n$ &$\beta_1$ &$P$ &token acc. &type acc.\\
\midrule
\textit{mean $\beta_1$}         &Bulgarian      &13018  &20.058 &0.000  &0.650  &0.649  \\
18.591 [15.740, 21.443]         &Chinese        &5909   &18.604 &0.000  &0.724  &0.766  \\
        &Croatian       &15555  &21.246 &0.000  &0.666  &0.634  \\
\textit{mean token accuracy}    &Czech  &27899  &28.207 &0.000  &0.671  &0.665  \\
0.656 [0.630, 0.683]    &Danish &11226  &17.506 &0.000  &0.786  &0.770  \\
        &Dutch  &11279  &12.201 &0.000  &0.609  &0.605  \\
\textit{mean type accuracy}     &English        &23311  &22.076 &0.000  &0.643  &0.647  \\
0.645 [0.616, 0.674]    &Finnish        &12605  &15.342 &0.000  &0.655  &0.644  \\
        &German &16391  &16.210 &0.000  &0.601  &0.606  \\
        &Greek  &5506   &18.383 &0.000  &0.631  &0.643  \\
        &Latvian        &5290   &15.826 &0.000  &0.594  &0.551  \\
        &Russian        &25397  &25.697 &0.000  &0.658  &0.651  \\
        &Slovak &11933  &25.935 &0.000  &0.700  &0.651  \\
        &Slovenian      &18859  &28.192 &0.000  &0.670  &0.661  \\
        &Swedish        &10937  &11.462 &0.000  &0.717  &0.711  \\
        &Turkish        &12115  &12.579 &0.000  &0.576  &0.577  \\
        &Ukrainian      &11474  &15.949 &0.000  &0.593  &0.592  \\
        &Urdu   &6432   &9.170  &0.000  &0.673  &0.593\\[0.5cm]
\textbf{ANA} &language &$n$ &$\beta_1$ &$P$ &token acc. &type acc.\\
\midrule
\textit{mean $\beta_1$}         &Basque &3322   &-9.623 &0.000  &0.703  &0.678  \\
31.313 [16.786, 45.841]         &Catalan        &3117   &45.135 &0.000  &0.818  &0.814  \\
        &Croatian       &4912   &-3.411 &0.106  &0.608  &0.604  \\
\textit{mean token accuracy}    &French &5673   &43.349 &0.000  &0.771  &0.756  \\
0.737 [0.674, 0.799]    &Galician       &5020   &68.290 &0.000  &0.805  &0.806  \\
        &Indonesian     &1521   &-2.462 &0.138  &0.543  &0.524  \\
\textit{mean type accuracy}     &Italian        &9484   &36.658 &0.000  &0.681  &0.698  \\
0.726 [0.665, 0.787]    &Persian        &2598   &43.242 &0.000  &0.794  &0.766  \\
        &Polish &13481  &24.873 &0.000  &0.684  &0.655  \\
        &Portuguese     &7580   &32.374 &0.000  &0.734  &0.725  \\
        &Romanian       &2426   &46.823 &0.000  &0.730  &0.739  \\
        &Spanish        &9212   &57.813 &0.000  &0.744  &0.738  \\
        &Vietnamese     &2636   &24.013 &0.000  &0.962  &0.931\\[0.5cm]
\textbf{NAA} &language &$n$ &$\beta_1$ &$P$ &token acc. &type acc.\\
\midrule
\textit{mean $\beta_1$}         &Arabic &11595  &4.595  &0.000  &0.693  &0.660  \\
4.140 [3.128, 5.152]            &Basque &4899   &1.957  &0.000  &0.626  &0.635  \\
        &Catalan        &2878   &5.024  &0.000  &0.710  &0.722  \\
\textit{mean token accuracy}    &French &8368   &5.143  &0.000  &0.737  &0.749  \\
0.680 [0.639, 0.721]    &Galician       &1334   &5.776  &0.000  &0.716  &0.694  \\
        &Hebrew &6751   &1.115  &0.000  &0.558  &0.560  \\
\textit{mean type accuracy}     &Indonesian     &5724   &4.631  &0.000  &0.740  &0.734  \\
0.687 [0.647, 0.726]    &Italian        &4523   &4.057  &0.000  &0.713  &0.739  \\
        &Persian        &12683  &1.583  &0.000  &0.605  &0.606  \\
        &Portuguese     &5139   &5.329  &0.000  &0.726  &0.730  \\
        &Romanian       &8492   &5.333  &0.000  &0.742  &0.746  \\
        &Spanish        &6245   &6.214  &0.000  &0.713  &0.745  \\
        &Vietnamese     &3354   &3.068  &0.000  &0.561  &0.606\\
\midrule
&\multicolumn{2}{r}{\textit{comprehensive mean}} &18.08 & &0.687 &0.681\\
\bottomrule
\end{tabular}
\caption{Results by template and language: $n$ triples analyzed, regression coefficient $\beta_1$ and $P$-value, and test accuracies. Means with 95\% confidence intervals shown for each template.}
\label{tab:results}
\end{table*}

As reported in Table \ref{tab:results}, we find above-chance $(>50\%)$ accuracy for all languages tested. We accurately predict 65.6\% of AAN triples, 73.7\% of ANA triples, and 68.0\% of NAA triples, for a comprehensive accuracy across all languages of 68.7\%. Overlapping 95\% confidence intervals across template means suggest that IG-based prediction performs equally well across templates.

Though we cannot make a direct comparison to other studies due to a lack of standardized datasets, our cross-linguistic accuracy of 68.4\% based on a single predictor compares reasonably favorably to a previous analysis of English AAN triples which achieved 72.3\% accuracy using a combination of predictors \citep{futrell2020what}.

The high performance on Vietnamese ANA triples (96.2\%) is largely due to the algorithm correctly predicting that the highly-frequent adjective \textit{{\fontencoding{T5}\selectfont nhiều}} `many' should be placed before the noun, while most other adjectives are placed after.\footnote{One might worry about the classification of `many' as an adjective. While widely extant across languages, the class of adjectives is not entirely homogeneous. As such, the equivalent of a word like `many' in some languages might be marked as an adjective, determiner, or other syntactic category. For the current study, we simply follow the UD annotation scheme.}

The learned $\beta_1$ coefficient is not significantly different between AAN (18.591) and NAA (31.313) triples, though that of NAA (4.140) triples is significantly smaller than the other two. More generally, of the 44 datasets tested, $\beta_1$ is positive in 41 (93.2\%), suggesting that there is a strong preference to maximize information gain. Further, of the three instances of a negative $\beta_1$, two (Croatian and Indonesian ANA) do not reach significance, perhaps due to a paucity of data. The sole significant negative $\beta_1$ is from Basque ANA triples.

\section{Discussion}

\subsection{Asymmetries}
The preference for one variant of an ANA triple over the other is an asymmetry without a straightforward explanation in a distance-based model; there is no clear mapping from ANA onto the other templates, which means that an adjective's relative distance to the noun is not informative. 
Our algorithm is novel in that the placement of the adjectives is governed by greedy IG, not distance to the noun---an innovation that allows us to break the symmetry between the adjectives in ANA triples. Similarly, IG makes no a priori prediction as to whether a mirror- or same-order will emerge between AAN and NAA triples: both pre- and post-nominal behavior is a product of ordering adjectives such that information gain is maximized, and IG itself is fundamentally derived from the distribution of adjectives and nouns that populate a language's possible feature vectors for conveying meaning.

Another left--right asymmetry that has been posited in the linguistics literature holds that dependents placed before the head in a surface realization (e.g., the adjectives in an AAN triple) follow a more rigid ordering than those placed after (e.g., the adjectives in a NAA triple; \citealp{hawkins1983word}). Both noun modifiers in general and adjectives specifically have been reported to follow this pattern, with a largely-universal pre-nominal ordering and a mirror, same, or `free' post-nominal order \citep{hetzron1978on}. However, there is as yet no large-scale empirical evidence for this claim.  

In an effort to empirically assess the claim that post-nominal orderings are more flexible compared to orderings pre-nominally, Table \ref{tab:reversed} reports the average prevalence of adjective pairs attested in both possible orders (e.g., A$_1$A$_2$N and A$_2$A$_1$N, where N can be any noun) within each template in our dataset. At 95\% confidence the difference between AAN and NAA does not reach significance, though the rate for ANA is significantly lower than the other two. More generally, the mean rate of just 1.6\% across templates reinforces the notion that ordering preferences are quite robust regardless of template, at least for our normalized triples from the languages analyzed here.

\begin{table}[tp]
\centering \setlength{\tabcolsep}{12pt}
\begin{tabular}[t]{@{} l r r r @{}}
\toprule
&$n$ &rate &confidence interval\\
\midrule
AAN &18 &0.017 &[0.012, 0.022]\\
ANA &13 &\textbf{0.007} &[0.002, 0.011]\\
NAA &13 &0.022 &[0.013, 0.032]\\
\midrule
all &44 &0.016 &[0.012, 0.020]\\
\bottomrule
\end{tabular}
\caption{Macro-average rate of adjectives attested in both orders by template, showing $n$ languages, rate, and 95\% confidence intervals.}
\label{tab:reversed}
\end{table}

\subsection{Ablation}
Equation \ref{eq:ig-kl} defines information gain as the conditioned sum of two elements, the positive evidence $D_{\text{KL}}[L^\prime || L]$ and the negative evidence $D_{\text{KL}}[\bar{L^\prime}||L]$. The positive evidence alone is akin to surprisal, a well-studied quantity in psycholinguistics (\ref{ssec:relationship}). By ablating the IG formulation into the two terms discretely, we can show empirically that the proportionally-combined positive and negative evidence yield more accurate and consistent results than either of the two constituent terms alone.

\begin{table*}\centering \setlength{\tabcolsep}{13pt}
\begin{tabular}[t]{@{} l r r r r r r r r@{}}
\toprule
&\multicolumn{4}{c}{accuracy} &\multicolumn{4}{c}{proportion of positive $\beta_1$}\\
\cmidrule(lr){2-5} \cmidrule(l){6-9}
 &AAN &ANA &NAA &all  &AAN &ANA &NAA &all\\
\midrule
$D_{\text{KL}}[L^\prime || L]$ &0.565 &0.567 &0.565 &0.566 &0.000 &0.154 &0.769 &0.273\\
$D_{\text{KL}}[\bar{L^\prime}||L]$ &0.533 &0.548 &0.526 &0.535 &0.167 &0.231 &0.462 &0.273\\
IG &\textbf{0.657} &\textbf{0.737} &\textbf{0.680} &\textbf{0.687} &\textbf{1.000} &\textbf{0.769} &\textbf{1.000} &\textbf{0.932}\\
\bottomrule
\end{tabular}
\caption{Ablation on accuracy and the proportion of positive coefficients for positive evidence ($D_{\text{KL}}[L^\prime || L]$), negative evidence ($D_{\text{KL}}[\bar{L^\prime}||L]$), and proportionally combined (IG). Boldfaced values indicate the highest accuracy or coefficient polarity proportion in each column.}
\label{tab:ablation}
\end{table*}

Table \ref{tab:ablation} shows the mean accuracy and polarity proportion of the $\beta_1$ coefficient across languages and templates. The polarity of $\beta_1$ tells us whether maximizing IG (positive) or minimizing IG (negative) is the better strategy. Thus a polarity percentage close to 0 or 1 indicates more consistent behavior across templates.

For example, while the accuracy of using only positive evidence, $D_{\text{KL}}[L^\prime || L]$, is 0.565, that accuracy is realized due to a 0.000 rate of positive $\beta_1$ coefficient---that is, the 56.5\% accuracy is achieved by minimizing IG, placing the adjective with the lower IG first. On the other hand, while using only positive evidence to predict NAA triples yields the same accuracy, 0.565, the coefficient polarity proportion of 0.769 means that in most NAA cases IG should be maximized. The three templates together reflect a modest accuracy (0.566) and an inconsistent coefficient polarity proportion (0.273).

Using only negative evidence, $D_{\text{KL}}[\bar{L^\prime}||L]$, yields even worse accuracies and similarly inconsistent coefficients as positive only. Accuracy across templates is little better than chance at 0.535, and the average coefficient polarity proportion of 0.273 likewise demonstrates that using negative evidence alone does not produce consistent behavior across templates.

The full IG calculation, including both positive and negative evidence, yields the highest accuracy across templates (0.687), as well as the highest for each template\textemdash AAN (0.657), ANA (0.737) and NAA (0.680). IG also demonstrates the most consistent behavior across languages and templates: at a rate of 0.932, maximizing IG yields the highest accuracy, regardless of whether adjectives precede or follow the noun.

\subsection{An efficient algorithm}\label{ssec:algorithm}
The goal of algorithms such as ID3 is to produce a decision tree which divides a dataset into equal-sized and mutually-exclusive partitions, thereby creating a shallow tree \citep{quinlan_induction_1986}. While finding the smallest possible binary decision tree is NP-complete \citep{hyafil1976constructing}, ID3's locally-optimal approach has proven quite effective at producing shallow trees capable of accurate classification \citep{dobkin1996induction}.

By analogy, the ordering of adjectives in a noun phrase by maximizing information gain likewise produces a tree with balanced positive and negative partitions at each node. Specifically, adjectives that minimize the entropy of both the positive and negative evidence are placed before adjectives which are less `decisive' at partitioning feature vectors.

\section{Summary}
We have taken a novel approach to the problem of predicting the surface order of adjectives across languages, casting it as a decision tree operating on a probability distribution over binary feature vectors. As each adjective is uttered, probability mass is partitioned into positive and negative subsets: those vectors which contain the feature and those that do not. The information gained by this partition can be used to order adjectives in a greedy manner, similarly to well-known algorithms for ordering nodes in a decision tree.

An IG-based approach allows us to provide the first quantitative information-theoretic account predicting the order of ANA triples. Further, with this approach we need not stipulate mirror- or same-orders for AAN and NAA triples. Because IG is not a distance metric between adjective and noun, and because IG incorporates negative evidence, both ANA and pre- or post-nominal asymmetries emerge within an IG framework, without appeal to other mechanisms.

Our results show that information gain is a good predictor of adjective order across languages. Importantly, IG-based prediction follows a consistent pattern across the three typological templates, namely that adjectives which maximize information gain tend to be placed first.

\bibliographystyle{acl_natbib}
\bibliography{acl}

\end{document}